%% file: ReID-Evaluation.tex
\documentclass[final]{cvpr}

\usepackage{times}
\usepackage{epsfig}
\usepackage{graphicx}
\usepackage{amsmath}
\usepackage{amssymb}

\usepackage{cite}
\usepackage{url}
\usepackage{booktabs}
\usepackage{makecell}
\usepackage{xcolor} 
\usepackage{multirow}
\usepackage{enumerate}
\usepackage{pifont}
\usepackage{array}   
\usepackage{caption}
\usepackage{algorithm, algorithmic}
\usepackage{verbatim}

\input{setup/macro}     
\input{setup/color}

\usepackage[pagebackref=true,breaklinks=true,colorlinks,bookmarks=false]{hyperref}

\newcounter{daggerfootnote}

\begin{document}

\title{Re-identification = Retrieval + Verification: \\
Back to Essence and Forward with a New Metric
}

\author{Zheng Wang$^{1,}$\thanks{Equal contributions. Zheng Wang is the corresponding author.} \hspace{0.1in} Xin Yuan$^{2,*}$ \hspace{0.1in} Toshihiko Yamasaki$^{1}$ \hspace{0.1in} Yutian Lin$^{3}$ \hspace{0.1in} Xin Xu$^{2}$ \hspace{0.1in} Wenjun Zeng$^{4}$ \hspace{0.1in}\\
$^{1}$The University of Tokyo \hspace{0.10in} $^{2}$Wuhan University of Science and Technology\\
$^{3}$Wuhan University \hspace{0.10in} $^{4}$Microsoft Research \\
{\tt\small wangz@hal.t.u-tokyo.ac.jp}
}
\maketitle

\begin{abstract}
Re-identification (re-ID) is currently investigated as a closed-world image retrieval task, and evaluated by retrieval based metrics. The algorithms return ranking lists to users, but cannot tell which images are the true target. In essence, current re-ID overemphasizes the importance of retrieval but underemphasizes that of verification, \textit{i.e.}, all returned images are considered as the target. On the other hand, re-ID should also include the scenario that the query identity does not appear in the gallery. To this end, we go back to the essence of re-ID, \textit{i.e.}, a combination of retrieval and verification in an open-set setting, and put forward a new metric, namely, Genuine Open-set re-ID Metric (GOM). 

GOM explicitly balances the effect of performing retrieval and verification into a single unified metric. It can also be decomposed into a family of sub-metrics, enabling a clear analysis of re-ID performance. We evaluate the effectiveness of GOM on the re-ID benchmarks, showing its ability to capture important aspects of re-ID performance that have not been taken into account by established metrics so far. Furthermore, we show GOM scores excellent in aligning with human visual evaluation of re-ID performance. Related codes are available at \textcolor{magenta}{\url{https://github.com/YuanXinCherry/Person-reID-Evaluation}.}
\end{abstract}

\begin{figure}[t]
	\centering
	\includegraphics[width=\columnwidth]{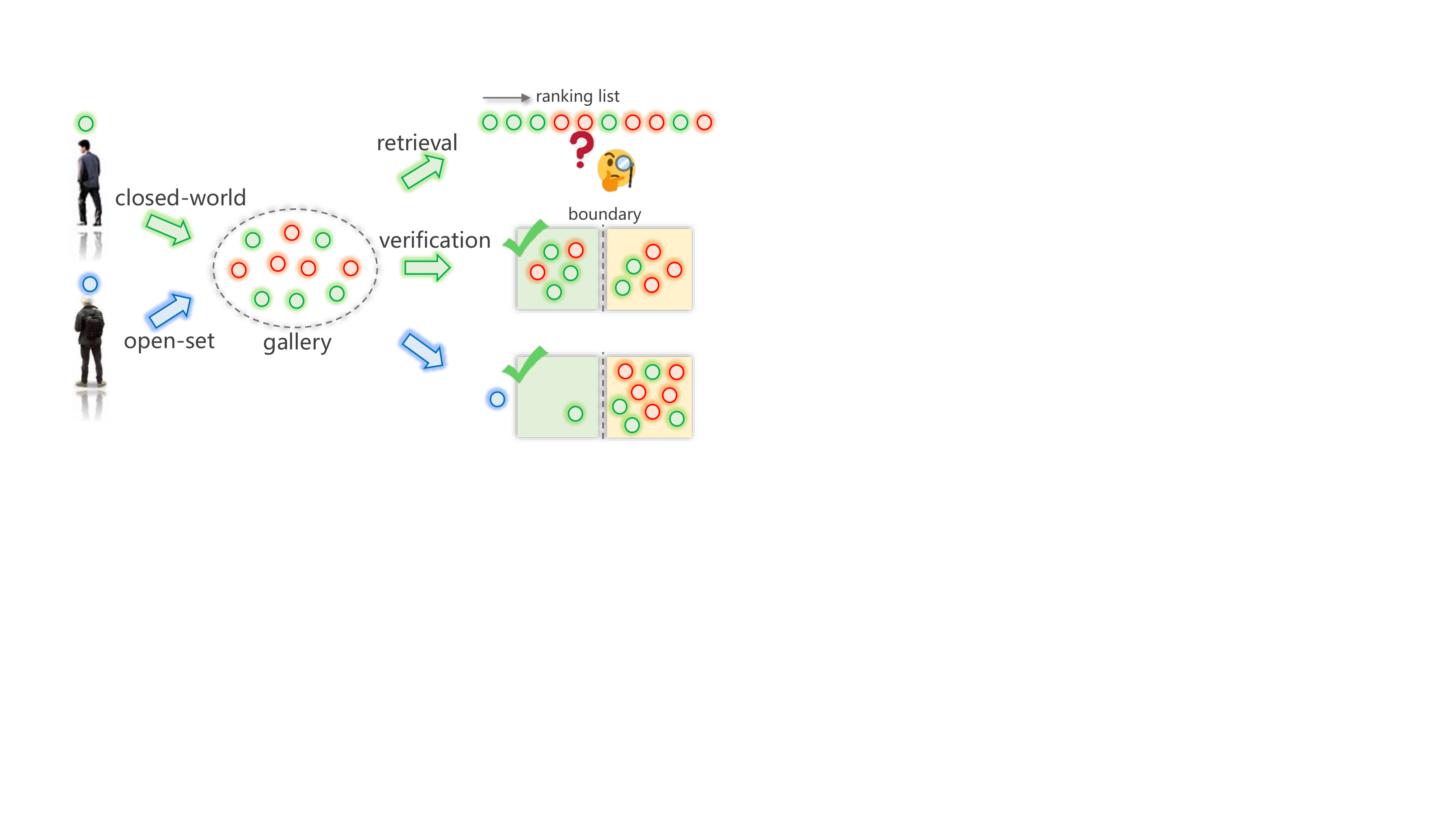}
	\vspace{-4mm}
    \caption{Illustration of existing retrieval and verification procedures. 1) Closed-world retrieval: the method returns a ranking list to users and cannot tell which images are the true targets. The user needs to judge targets according to their experience and feeling. 2) Closed-world verification: given a decision boundary, images whose distances are below the boundary are considered as the targets. Nevertheless, the method cannot distinguish ground truth (GT) and non-GT within the boundary. 3) Open-set: the GT of the probe does not always exist in the gallery, thus re-ID procedure should include this kind of scenario.}
    \vspace{-3mm}
\label{fig:intro_motivation}
\end{figure}

\section{Introduction}
Re-identification (re-ID) is the task of finding out the presence of person/vehicle-of-interest in a database of images captured from multiple different cameras in a wide-area cross-camera video surveillance scenario~\cite{wang2016zero}. The re-ID task is one of the growing communities of computer vision research, and is imperative for the intelligent video surveillance system. Yet, the evaluation of almost all re-ID algorithms is done under a \textit{\textbf{closed-world retrieval}} setting. Under this setting, existing evaluation metrics can be classified into two types according to the number of ground truth (GT) that exists for each query. The first type is a single-GT evaluation with Cumulative Matching Characteristics (CMC)~\cite{gray2007evaluating}. The second type is a multi-GT evaluation with mean Average Precision (mAP)~\cite{zheng2015scalable}. Also, a novel metric named mean inverse negative penalty (mINP)~\cite{ye2020deep}, as a supplement, has been proposed to evaluate the property of the hardest correct match. However, existing retrieval based metrics are not able to reveal which algorithms are outstanding in the scenario that the system needs to output some targets automatically rather than a ranking list with no clear targets. Users still need to find out GTs from returned ranking lists via their eyes (\textit{retrieval} in Figure~\ref{fig:intro_motivation}). In essence, the procedure of re-ID expects machines to output targets automatically without human involving, so that further tracking, behaviour analysis, and investigation will continue smoothly. 

To our knowledge, few methods~\cite{li2013learning,zhengweishi2016tpami,kushwaha2020pug} studied and formulated re-ID as a \textit{\textbf{verification}} task. Similar to the detection and classification tasks, where, given a threshold, the algorithms can be evaluated by the metrics of precision~\cite{li2013learning} and recall~\cite{kushwaha2020pug} on the automatically predicted targets. However, the verification based metrics are not able to indicate which returned targets are GTs and
which GTs are not returned (\textit{verification} Figure~\ref{fig:intro_motivation}). In practice, it is still required that the true positive ranks top in predicted targets. Thus, we consider that re-ID cannot be formulated as retrieval or verification alone. It is essentially a combination of retrieval and verification. In this paper, we propose to design new metrics to benefit the re-ID community with a comprehensive evaluation and making it forward in its right way.

In addition, current re-ID methods have made an unscalable \textit{\textbf{closed-world}} assumption, where every probe identity must guarantee to appear in the gallery set. Nevertheless, the GT of the probe does not always exist in the gallery in real-world applications. Typically, it is frequent to search for an inconclusive suspect who may not be in a video or a large set of videos in realistic forensic video analysis. In this case, it is called an \textit{\textbf{open-set}} problem (\textit{open-set} Figure~\ref{fig:intro_motivation}), which has been less investigated~\cite{liao2014open,zhengweishi2016tpami,zhu2018tip}. Specifically, the false accept rate (FAR)~\cite{liao2014open} and the false target rate (FTR)~\cite{zhengweishi2016tpami} were designed to tell the proportion of non-target probes verified as the target. They are also often reported with verification previsions by the Receiver Operating Characteristic (ROC)~\cite{liao2014open} under different decision thresholds. However, open-set based metrics above mainly focus on the probes that do not have GTs in the gallery, not comprehensively evaluating the performance of re-ID algorithms.

\tabcolsep=3.5pt
\begin{table}
\centering
\caption{Comparison with existing metrics.}
\resizebox{\linewidth}{!}{
\begin{tabular}{l|ccccc} 
\toprule
\multirow{2}{*}{\textbf{Metric}}  & \multirow{2}{*}{\textbf{Closed/Open}} & \multirow{2}{*}{\textbf{Task}}  & \multicolumn{3}{c}{\textbf{\#GT in Gallery}} \\
 & & & Single & Multiple & None  \\
\midrule
\midrule
CMC~\cite{gray2007evaluating}                          & closed-world & Retrieval      & \checkmark &            &              \\
mAP~\cite{zheng2015scalable} / mINP~\cite{ye2020deep}    & closed-world & Retrieval     & \checkmark & \checkmark &              \\
precision~\cite{li2013learning} / recall~\cite{kushwaha2020pug}  & closed-world  & Verification   & \checkmark & \checkmark & \\
FAR~\cite{liao2014open} / FTR~\cite{zhengweishi2016tpami} & open-set   & Verification   & \checkmark  &            & \checkmark  \\
\midrule
GOM                                                  & open-set     & re-ID           & \checkmark & \checkmark & \checkmark  \\
\bottomrule
\end{tabular}
}
\label{tab:metrics}
\end{table}

To summarize the disparities between our novel metrics and existing metrics, a comparison is illustrated in Table~\ref{tab:metrics}. To remedy the above problems of existing metrics (\ie not really re-ID procedure), we present a novel \underline{G}enuine \underline{O}pen-set re-ID \underline{M}etric (GOM) in this paper, including mean re-ID Precision curve ($mReP$) and mean False Rate curve ($mFR$) respectively. $mReP$ and $mFR$ together are capable of evaluating genuine re-ID in both closed-world and open-set, promoting machines to automatically select and return target images. We will provide analysis as to why GOM is preferable to current alternatives for evaluating re-ID algorithms. In essence, current re-ID overemphasizes the importance of retrieval but underemphasizes that of verification as well as the open-set problem. GOM explicitly measures both properties of the two tasks, and combines these in a balanced way. 

The GOM metric is also intuitive to understand (please refer to Table~\ref{tab:example} and its illustration in Section~\ref{Sec: analysis of GOM}). It consists of several sub-metrics. The Retrieval Precision $RP$ is similar to AP. The Verification Precision $VP$ is simply the percentage of returned GTs between GTs and returned negatives. The $ReP$ score is the geometric mean of $RP$ and $VP$ over different decision thresholds. The false rate $FR$ is simply the percentage of false targets. The final two metrics are $mReP$, averaged over all $ReP$, and $mFR$, averaged over all $FR$. We make three significant novel contributions:

1) We propose a novel metric GOM for evaluating genuine open-set re-ID. GOM can be decomposed into a family of sub-metrics which are able to evaluate different aspects of re-ID separately. This enables a clear understanding of the different types of tasks that the feedback rank lists are returned based on threshold, and enables decision thresholds to be tuned for different requirements.
 
2) We investigate a thorough theoretical analysis of GOM as well as commonly used metrics CMC~\cite{gray2007evaluating}, mAP~\cite{zheng2015scalable}/mINP~\cite{ye2020deep}, precision~\cite{li2013learning}/recall~\cite{kushwaha2020pug}, and  FAR~\cite{liao2014open}/FTR~\cite{zhengweishi2016tpami}, emphasizing the benefits and drawbacks of each metric. 

3) We evaluate GOM on two person re-ID datasets (Market-1501~\cite{zheng2015scalable} and DukeMTMC-reID~\cite{Zheng_2017_ICCV}) and one vehicle re-ID dataset (VeRi776~\cite{Liu_2016_ECCV}), and analyze its properties compared to other metrics for evaluating current state-of-the-art re-ID methods. Besides, we have exhibited a user study comparing how different metrics align with the human judgment of re-ID and show that GOM well aligns with the desired re-ID's procedure of users.

\section{Related Work}
\heading{History of re-ID metrics.} The re-ID task originates from multi-camera tracking~\cite{wang2013prl}. In the early years, re-ID datasets were collected by two cameras, such as VIPeR~\cite{Douglaseccv2008}, and each query had only one GT. Hence, early works tended to evaluate using the Cumulative Matching Characteristic (CMC)~\cite{gray2007evaluating} curve. It has often been highly criticized~\cite{zheng2015scalable} for its properties that focus on the most similar GT. Recently, large scale datasets, such as Market-1501~\cite{zheng2015scalable} and DukeMTMC-reID~\cite{Zheng_2017_ICCV}, have been constructed, which contain multiple GTs. To improve the limitations of CMC, Zheng \etal~\cite{zheng2015scalable} proposed to use the mean average precision (mAP) for evaluation, which is originally widely used in image retrieval. It can address the issue of two systems performing equally well in searching the first GT, but having different retrieval abilities for other hard GTs. Recently, mINP~\cite{ye2020deep} was proposed to evaluate the ability to retrieve the hardest GT, which avoids the domination of easy matches in the mAP/CMC evaluation. On the other hand, the precision and recall metrics, specifically proposed for detection and classification tasks, have been used by few re-ID methods~\cite{li2013learning,kushwaha2020pug}, due to its focus on measuring the verification accuracy rather than the retrieval accuracy. However, they exhibit unintuitive and atypical in regards to retrieval.

\heading{Metrics of open-set re-ID.} Closed-world re-ID aims to determine which gallery image is the probe. However, numerous irrelevant identities exist, and the probe identity may not appear in the gallery, \ie open-set re-ID. Open-set re-ID is usually formulated as a verification problem~\cite{liao2014open, zhengweishi2016tpami,zhu2018tip}. Zheng \etal~\cite{zheng2012tansfersreid} conducted the first open-set re-ID work. Liao \etal~\cite{liao2014open} decomposed open-set re-ID into detection and identification, where identification rate (DIR) and false accept rate (FAR) were discussed. Zheng \etal~\cite{zhengweishi2016tpami} utilized the true target rate (TTR/FTR similar to DIR/FAR) for performance evaluation. Note that the Receiver Operating Characteristic (ROC)~\cite{liao2014open} is made up of DIR and FAR. Although the open-set re-ID is much closer to practical video surveillance applications than the closed-world, the attention devoted to this issue is relatively limited for three possible reasons. First, closed-world re-ID is a utilizable technical route, and it is convenient and fair for conducting research on previous theories and technologies due to various baselines and datasets. Second, the low recognition rates under low false accepted rates of existing methods show that open-set re-ID is challenging. Third, there are no public datasets designed for open-set re-ID evaluation.

\heading{Multi-Target Multi-Camera Tracking (MTMCT).} MTMCT aims to determine the cross-camera trajectories of certain targets captured from multiple cameras~\cite{chen2020tcsvt}. MTMCT and open-set re-ID could fall under the cross-camera identity verification task. Recently, Ristami \etal~\cite{Ristani_2018_CVPR} presented a CNN feature for both MTMCT and re-ID. However, MTMCT still has several differences from re-ID. The former focuses on reducing classification error rates, whereas the latter pays attention to improving the ranking performance at a certain error-tolerant rate.

\heading{Meta-Evaluation of metrics.} Vasant \etal~\cite{manoha2006petsvsvace} investigated the trade-off between providing multiple evaluation metrics versus a single unifying metric. Compared to multiple metrics which are helpful for researchers or developers to debug algorithm, a unified metric can be useful to assist end-users to easily choose effective models from various options. Existing re-ID metrics have different views, but are designed for the retrieval task or verification task, essentially, not suitable as a unified metric for genuine re-ID. Srikrishna \etal~\cite{Srikrishna2019pami} also pointed out that researchers should take a broader view of how algorithms perform and not just look at raw rank-1 or mAP numbers. We present GOM as a suitable solution for unified comparison.
 
\section{Preliminaries}

This section describes the re-ID task, the role of evaluation metrics, as well as the definition of open-set re-ID.

\heading{What is re-ID?} Re-ID is one of the core tasks for the intelligent video surveillance system. The input is a query/probe that came from the human need. The output is a batch rank list of gallery based on distance/similarity that contains the information about: 1) whether this target exists or appears in the gallery (open-set)? 2) whether all targets are found out in the ranking list (verification)? 3) which images belong to the same identity (retrieval).

\heading{Evaluation metrics.} The choice of evaluation metric is extremely important, as the properties of the metric determine how different errors contribute to a final score. The choice of metric also has the ability to heavily influence the direction of re-ID research. In the age of competitive benchmarks, a lot of researches are evaluated on the ranking ability on the benchmarks. This will also guide researches and methods towards focusing on these aspects.

\heading{Definition of open-set re-ID.} Suppose that there are two query sets $\mathcal{Q}_+$ and $\mathcal{Q}_-$. $\mathcal{Q}_+$ consists of $N$ images that have the GTs in the gallery set $\mathcal{G}$, and $\mathcal{Q}_-$ consists of $M$ images that do not have the GTs in $\mathcal{G}$.

\section{Overview of Previous Metrics}
A brief revisit of the current commonly used evaluation metrics is as follows:

\subsection{CMC}
For each query $i \in \mathcal{Q}_+$, all gallery images $k \in \mathcal{G}$ are ranked based on the distance $d(i,k)$. Cumulative Match Characteristic (CMC)~\cite{gray2007evaluating} counts if top-$x$ ranked gallery samples contain the query identity. The rank index of the correct match is denoted as $R^{k^{*}}_i$. The gallery image $k^{*}$ has the same identity as the query $i$. $CMC@x$ is defined as: 
\begin{equation}
    CMC@x = \frac{\sum _{i=1}^{N}\mathbb{I}(R^{k^{*}}_i \leq x)}{N},
\end{equation}
where $\mathbb{I}(\cdot)$ is the measure function. Note that CMC does not have a common agreement when it comes to the multi-gallery-shot setting, where each gallery identity could have multiple instances.

\subsection{mAP}
For each query $i \in \mathcal{Q}_+$, an average precision (AP)~\cite{zheng2015scalable} calculates the area under the Precision-Recall curve $AP_i = \sum _x prec_i^x(recall_i^x - recall_i^{x-1})$, where $prec_i^x = \frac{|\{k | k \in G_i , R^{k}_i\leq x\}|}{|x|}$ and $recall_i^x = \frac{|\{k | k \in \mathcal{G}_i , R^{k}_i\leq x\}|}{|\{k | k \in \mathcal{G}_i\}|}$. $\mathcal{G}_i$ stands for the sub gallery set that have the same person ID as the query $i$. Then, mAP evaluates the overall performance:
\begin{equation}
    mAP = \frac{1}{N} \sum _{i=1}^{N} AP_i.
\end{equation}

\subsection{mINP}
The negative penalty (NP)~\cite{ye2020deep} measures the penalty to find the hardest correct match $NP_i = \frac{R^{hard}_i - |\mathcal{G}_i|}{R^{hard}_i}$, where $R^{hard}_i$ indicates the rank position of the hardest match, and $|\mathcal{G}_i|$ represents the total number of correct matches for query $i$. For consistency with CMC and mAP, the inverse negative penalty (INP) is preferred. Overall, the mean INP (mINP) of all the queries is represented by: 
\begin{equation}
    mINP = \frac{1}{N}\sum _{i=1}^{N} (1-NP_i) = \frac{1}{N}\sum _{i=1}^{N}\frac{|\mathcal{G}_i|}{R^{hard}_i}.
\end{equation}

\subsection{FAR and ROC}
Liao \etal~\cite{liao2014open} considers the open-set re-ID problem. It defines an identification rate (DIR) $DIR(\tau, x) = \frac{|\{i | i \in \mathcal{Q}_+, R_i\leq x, d(i,k^{*}) \leq \tau\}|}{|\mathcal{Q}_+|}$ and a false accept rate (FAR) $FAR(\tau) = \frac{|\{i | i \in \mathcal{Q}_-, \min d(i,k) \leq \tau\}|}{|\mathcal{Q}_-|}$, where $\tau$ is the decision threshold, $x$ is the rank index. Receiver Operating Characteristic (ROC) curve considers a situation that the query is not present in the gallery. Given the rank $x$ and the decision threshold $\tau$, an ROC curve can be drawn by plotting DIR versus FAR.

\section{The GOM Evaluation Metric}
\label{sec:gom}
The main contribution of this paper is a novel evaluation metric for evaluating re-ID performance. We term this evaluation metric GOM (Genuine Open-set re-ID Metric). GOM builds upon some previously used metrics, while addressing many of their deficits. GOM is designed to: 1) provide scores for re-ID evaluation which combines different aspects of re-ID evaluation, 2) evaluate all over stage performance under different decision threshold, and finally, 3) decompose into sub-metrics which allow an analysis of the different components of retriever's or verifier's performance.

\heading{Retrieval Precision.} For a particular decision threshold $\tau$, we use $AP$ to record the retrieval precision of each query $i$:
\begin{equation}
    RP_{\tau,i} =  AP_{\tau,i}.
    \label{eq:RP}
\end{equation}

\heading{Verification Precision.} Also, given a decision threshold $\tau$ and a query $i$, the verification precision can be calculated as: 
\begin{equation}
    VP_{\tau,i} =  \frac{|\mathcal{G}^{TP}_{\tau,i}|}{|\mathcal{G}^{TP}_{\tau,i}|+|\mathcal{G}^{FN}_{\tau,i}|+|\mathcal{G}^{FP}_{\tau,i}|},
    \label{eq:VP}
\end{equation}
where $|\mathcal{G}^{TP}_{\tau,i}|$, $|\mathcal{G}^{FN}_{\tau,i}|$, and $|\mathcal{G}^{FP}_{\tau,i}|$ represent the number of correct match subset (true positive), the number of incorrect match subset (false positive), and the number of targets not be found (false negative) in the returned subset under the decision threshold $\tau$.

\heading{Re-ID Precision.} Then, the genuine re-ID precision $mReP$ evaluates the mean of the geometric mean of the retrieval precision and the verification precision by Eq.~\ref{eq:ReP}. This formulation ensures that both retrieval and verification are balanced. Note that it evaluates the overall performance of all queries that consist of targets in the gallery.
\begin{equation}
    mReP_{\tau} =  \frac {1}{N} \sum_{i = 1}^N  \sqrt{VP_{\tau,i} * RP_{\tau,i}}.
    \label{eq:ReP}
\end{equation}

\heading{False Rate.} We evaluate the false rate for all queries $\mathcal{Q}_-$ with no GTs in the gallery. For a query $j$, $FR_{\tau,j}$ calculates the proportion of incorrect matches:
\begin{equation}
    FR_{\tau,j} =  \min(\frac{|\mathcal{G}^{FP}_{\tau,i}|}{B},1),
\end{equation}
where $B$ is an integer constant\footnote{we set $B=3000$ when evaluating on Market-1501 and DukeMTMC-reID datasets.} to decide the false rate. Then, $mFR$ is used to evaluates the overall performance of all queries that do not consist of GTs in the gallery, as Eq.~\ref{eq:FR} represents.
\begin{equation}
    mFR_{\tau} = \frac {1}{M} \sum_{j = 1}^M \min(\frac{|\mathcal{G}^{FP}_{\tau,i}|}{B},1).
    \label{eq:FR}
\end{equation}
    
\heading{Integrating over decision thresholds.} According to different $\tau$, we can draw an $mReP$ curve and an $mFR$ curve. The re-ID precision and false rate scores are the integral of $mReP$ and $mFR$ scores across the valid range of $\tau$ values between 0 and 1. The results are calculated via the following formulation\footnote{$\tau$ values 0 to 1 with 0.01 intervals.}:
\begin{equation}
    MREP = \int_{0}^{1} mReP_{\tau} d{\tau} 
    \label{eq:MREP}
\end{equation}
\begin{equation}
    MFR = \int_{0}^{1} mFR_{\tau} d{\tau} 
    \label{eq:MFR}
\end{equation}

The evaluation process has three steps: 1) Given a distance matrix ($|\mathcal{Q}|\times |\mathcal{G}|$) generated by the evaluated method, we normalize it to $[0,1]$ and set a series of thresholds. 2) We calculate the $mReP_{\tau}$ and $mFR_{\tau}$ at such given threshold ${\tau}$ by Eq.~\ref{eq:ReP} and Eq.~\ref{eq:FR}. Then, we draw the $mReP$ and $mFR$ curves. 3) The $MREP$ and $MFR$ under different thresholds are calculated, and the important thresholds $\tau_{max}$ for $mReP_{\tau}$ and $\tau_{nz}$ for $mFR_{\tau}$ are recorded as well.

\section{Analysis of GOM}
\label{Sec: analysis of GOM}
In this section, we make an analysis of GOM in terms of taxonomy of error types, and use an example to show the advantage of our metrics. 

\heading{Analysis of re-ID errors.} We classify potential re-ID errors into three categories: retrieval errors, verification errors, and decision errors. 1) Retrieval errors occur when negative samples rank higher than positive samples in the gallery. We always use the metrics of CMC~\cite{gray2007evaluating}, mAP~\cite{zheng2015scalable}, and mINP~\cite{ye2020deep} to evaluate different methods. These retrieval based metrics all focus on the rank of samples in the whole gallery while ignoring which the real targets are. 2) Verification errors occur when the negative samples are in the returned results, or when the positive samples are not in the returned results. We often use the metrics of precision and recall to evaluate different methods. Similar to detection/classification evaluation, verification based metrics only focus on the proportion of returned/missed positives, but cannot tell the exact correct matches. 3) Decision errors occur on the condition that methods obtain returned samples when the query identity does not exist in the gallery set. FAR/FTR is the most representative metrics in the open-set setting. These evaluation metrics only reflect whether the query identity appears in the gallery, but are impossible to measure the correct match of returned results. 

\begin{figure}[t]
	\centering
	\includegraphics[width=0.85\columnwidth]{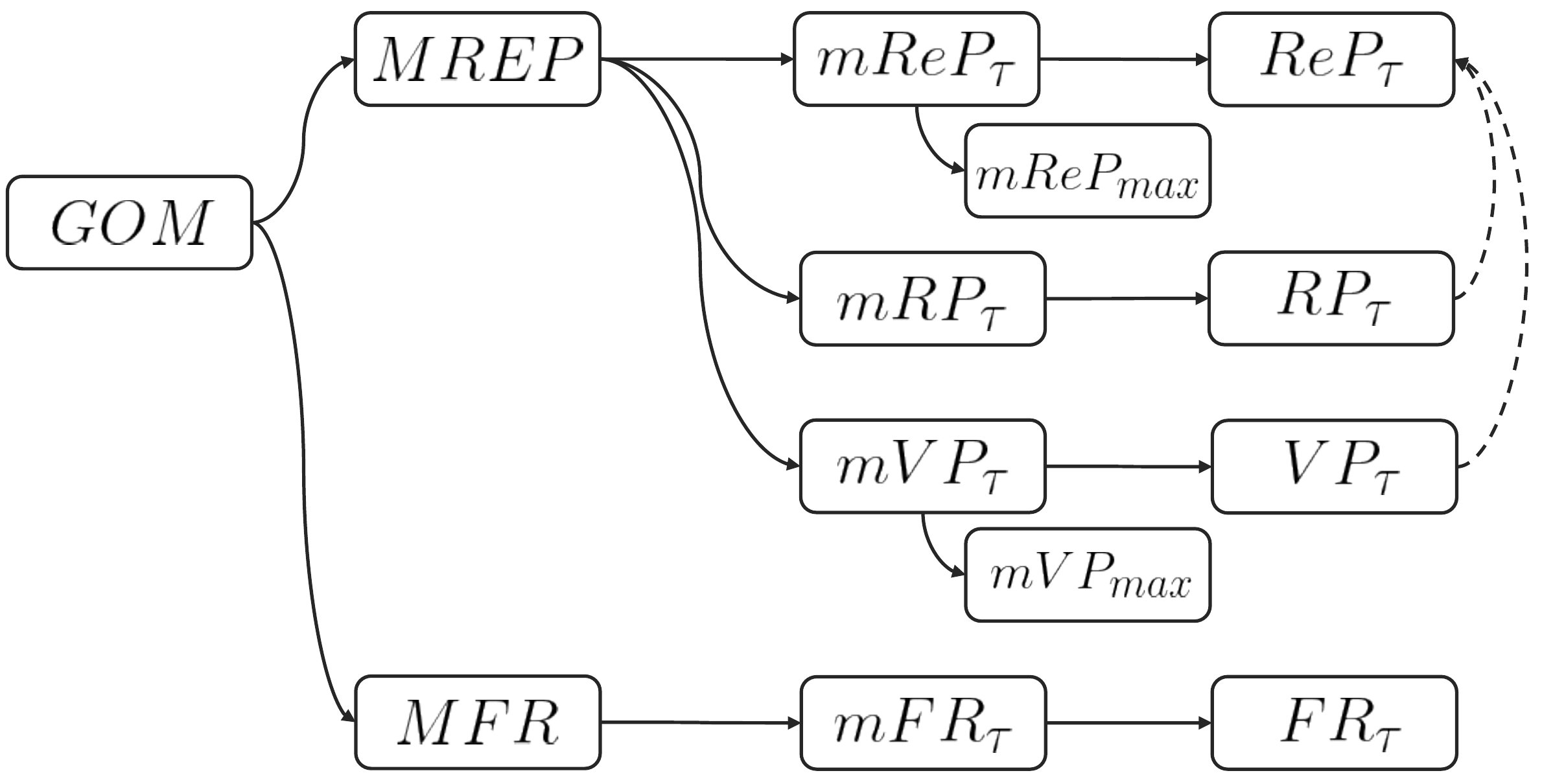}
    \caption{A family of metrics of GOM. Note that $mReP_{\tau}$, $mRP_{\tau}$, $mVP_{\tau}$ and $mFR_{\tau}$ stand for the mean value of $ReP_{\tau}$, $RP_{\tau}$, $VP_{\tau}$ and $FR_{\tau}$ scores by all queries, respectively. $ReP_{\tau}$ is calculated by the geometric mean of $RP_{\tau}$ and $VP_{\tau}$. $mReP_{max}$ and $mVP_{max}$ stand for the max value of $mReP_{\tau}$ and $mVP_{\tau}$ under all thresholds $\tau$.}
    \label{fig:metric}
\end{figure}

To better assess these three types of errors, we propose the novel evaluation metric GOM (see in Sec.~\ref{sec:gom}). Figure~\ref{fig:metric} shows the set of sub-metrics of GOM and their relations. The sub-metrics have two main purposes. The first purpose is to enable a simple comparison between methods to determine which performs better than the others. The second purpose of evaluation metrics is to enable the analysis of different types of errors, as well as the understanding of the different types of requirements. For the retrieval errors, we improve original average precision AP by adding a threshold $\tau$ and design the $RP_{\tau}$ (Eq.~\ref{eq:RP}) to compute retrieval accuracy in terms of thresholds. For the verification errors, to balance existing $precision$ and $recall$, we propose the verification precision $VP_{\tau}$ (Eq.~\ref{eq:VP}), similar to the standard Jaccard index that is commonly used for evaluating classification. Note that $VP_{\tau}$ can be represented by $precision$ and $recall$. Please refer to the supplementary for its detail. We can also consider both retrieval and verification errors, then combine $RP_{\tau}$ and $VP_{\tau}$ together using geometric mean $ReP_{\tau}$ (Eq.~\ref{eq:ReP}) under the same threshold. For the decision errors, we also propose a false rate $FR_{\tau}$ (Eq.~\ref{eq:FR}) in terms of the threshold, which shows whether fewer negatives are returned, while the existing metric FAR/FTR does not have this kind of ability.

\heading{Why geometric mean for \textit{\textbf{ReP}}.} The $ReP_{\tau}$ formulation contains a square root operation after the multiplication of $VP_{\tau}$ and $RP_{\tau}$. This square root has three effects. 1) The scores of the model can be increased in terms of the magnitude and spread. It is nice to observe that both the magnitude and spread of $ReP_{\tau}$ scores fall in the same range as previous metrics. This indicates that researchers’ current intuitive understanding of how good certain scores are still roughly holds. 2) $ReP_{\tau}$ can evenly balance both retrieval and verification via calculating the geometric mean of these two scores. The geometric mean is more suitable in re-ID than other formulations such as the arithmetic mean. When a model completely fails in either retrieval or verification, the score of the geometric mean can approach 0 to effectively represent this. 3) Double counting of similar error types can be taken into consideration. As discussed above, retrieval/verification errors are included in the $ReP_{\tau}$ score. These errors can be prevented via the use of the square root.

\tabcolsep=1pt
\begin{table*}
\centering
\begin{minipage}[c]{6.5cm}
  \includegraphics[width=\linewidth]{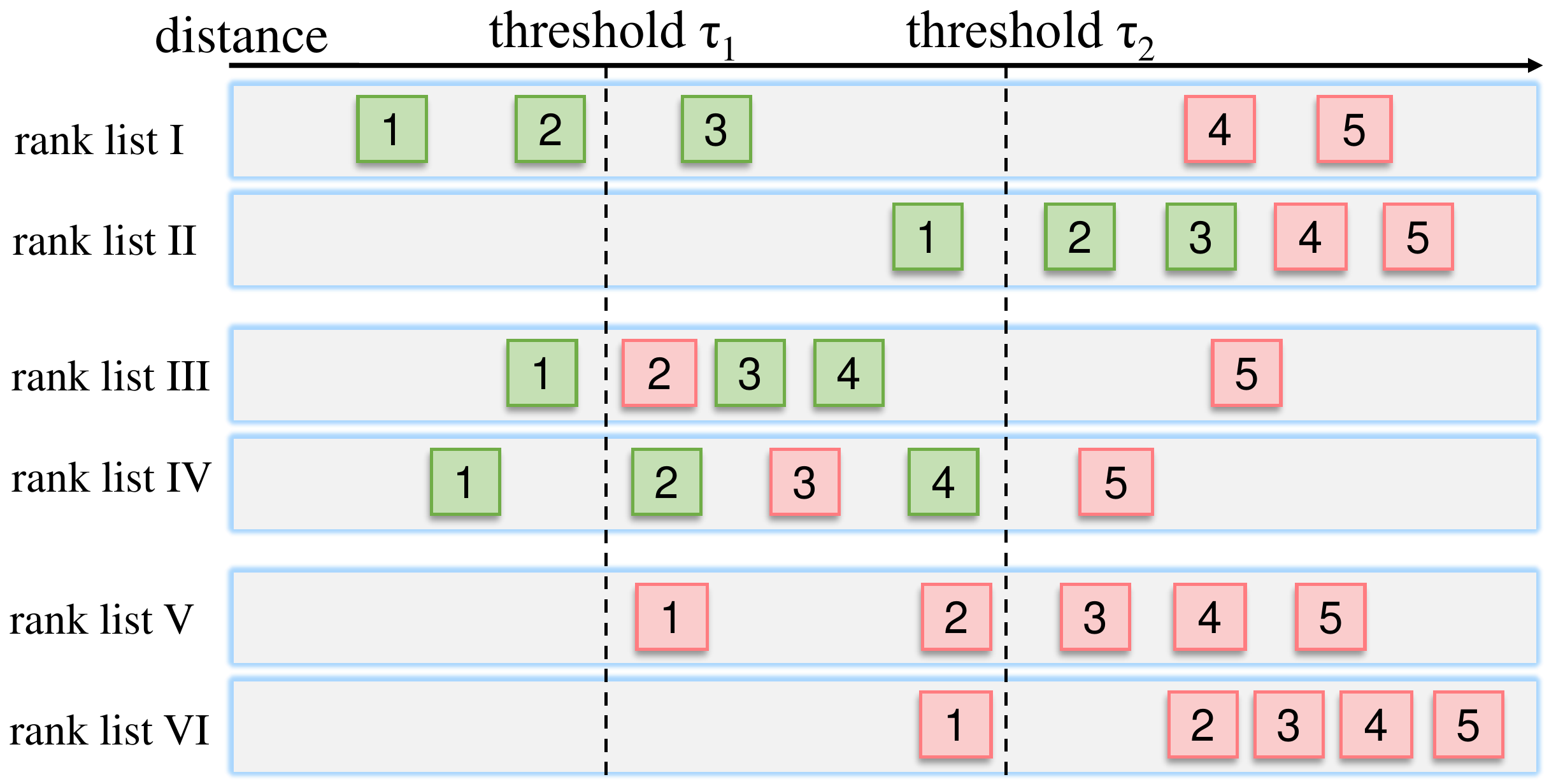}
\end{minipage}
\tabcolsep=1.5pt
\resizebox{0.56\linewidth}{!}{
\begin{tabular}{c|ccccc | cccc | cccc } 
\toprule
& \multicolumn{5}{c|}{Existing} & \multicolumn{8}{c}{Proposed}\\
           \cline{2-14} 
           \centering
           & CMC@1 & AP & INP & FAR$_{\tau _{1}}$ & FAR$_{\tau _{2}}$
           & $RP_{\tau _{1}}$ & $VP_{\tau _{1}}$ & $ReP_{\tau _{1}}$ & $FR_{\tau _{1}}\downarrow$  
           & $RP_{\tau _{2}}$ & $VP_{\tau _{2}}$ & $ReP_{\tau _{2}}$ & $FR_{\tau _{2}}\downarrow$ \\
\midrule
I     & 1     & 1   & 1  & --  & --  & 1 & 0.67  & 0.82 & -- & 1     & 1   & 1 & --\\
II    & 1     & 1   & 1  & --  & --  & 0 & 0     & 0    & -- & 1     & 0.33   & 0.57 & --\\
\midrule
III   & 1     & 0.81      & 0.75     & --  & --  & 1 & 0.33  & 0.57 & -- & 0.81  & 0.75   & 0.78    & --\\
IV    & 1     & 0.92      & 0.75     & --  & --  & 1 & 0.33  & 0.57 & -- & 0.92  & 0.75   & 0.83    & --\\
\midrule
V     & --    & --     & --    & \multicolumn{1}{c}{\multirow{2}{*}{0}}   & \multicolumn{1}{c|}{\multirow{2}{*}{1}} &-- & --    & --   & 0 & --    & --     & --   & 0.4\\
VI     & --    & --     & --    &     &   &-- & --    & --   & 0 & --    & --     & --   & 0.2\\
\bottomrule
\end{tabular}
}
\vspace{-2mm}
\caption{A simple re-ID example highlighting the main differences between evaluation metrics. Six different ranking lists are shown. $AP$ and $FAR$ overemphasize the task of retrieval and verification respectively. GOM balances both of these by being an explicit combination of a retrieval precision $RP$ and a verification precision $VP$, contributed to re-ID precision $ReP$, and assess the false rate under the different threshold by $FR$.}
\label{tab:example}
\end{table*}

\heading{Comparison with existing metrics using toy samples.} Table~\ref{tab:example} shows some toy samples and their results under existing metrics CMC@1, AP, INP, and FAR, and our metrics $RP_\tau$, $VP_\tau$, $ReP_\tau$, and $FR_\tau$. The examples demonstrate the following conclusions:

\begin{itemize}
\item For the rank list I and II, the figure shows that their retrieval results are the same. CMC, AP, and INP metrics also give the same scores. Actually, the verification results are different. For example, under threshold $\tau_1$, list I outputs two targets, while list II does not output any targets. This example shows that \textit{the results may have the same retrieval score, but different verification scores}. In this case, existing metrics designed for evaluating retrieval (CMC, AP, and INP) are ineffective. Our metric $VP_\tau$ shows the verification performance under different thresholds. $VP_\tau$ clearly reflects that the results of the rank list I (0.67 at $\tau_1$ and 1 at $\tau_2$) are better than that of rank list II (0 at $\tau_1$ and 0.33 at $\tau_2$). Furthermore, in order to comprehensively consider the retrieval and verification, we proposed evaluation metric $ReP_\tau$ by making a geometric mean of $AP_\tau$ and $VP_\tau$, it also demonstrates that the rank list I is preferable, even though existing metrics cannot distinguish them. 

\item For the rank list III and IV, the figure shows that their verification results are the same under thresholds $\tau_1$ (both return one target) and $\tau_2$ (both return three positives and one negative). Existing metrics (CMC, AP, and INP) cannot reflect this situation, but our metric $VP_\tau$ demonstrates that their verification scores are the same (0.33 at $\tau_1$ and 0.75 at $\tau_2$). These two lists are indeed different when evaluating with retrieval metrics. Our metric $RP_\tau$ at $\tau_2$ shows that the result of rank list IV (0.92) is better than that of rank list III (0.81). This example shows that \textit{the results may have the same verification score, but different retrieval scores}. Furthermore, the metric $ReP_\tau$ also demonstrates that the rank list IV is preferable, even though existing metrics and the verification metric cannot distinguish them. 

\item For the rank list V and VI, they demonstrate the situation of open-set re-ID. The figure shows that the rank list VI performs relatively better than the rank list V, because more negatives are in a larger distance. However, if we use existing metrics FAR$_\tau$/FTR to measure the results. The rank list V and VI get the same score (0 at $\tau_1$ and 0 at $\tau_2$). On the other hand, our metric $FR_\tau$ at $\tau_2$ clearly reflects that the result of rank list VI (0.2) is better than that of rank list V (0.4).
\end{itemize}

To sum up, currently used metrics overemphasize the importance of retrieval but underemphasize that of verification, while our set of metrics can perform comprehensive with retrieval, verification, and open-set tasks. GOM decomposes into a family of sub-metrics that are able to evaluate different aspects of re-ID separately, and it also enables a clear understanding of different aspects of re-ID under different thresholds.

\tabcolsep=0.1pt	
\begin{figure*}
	\begin{tabular}{cccc}
		\includegraphics[width=.249\textwidth]{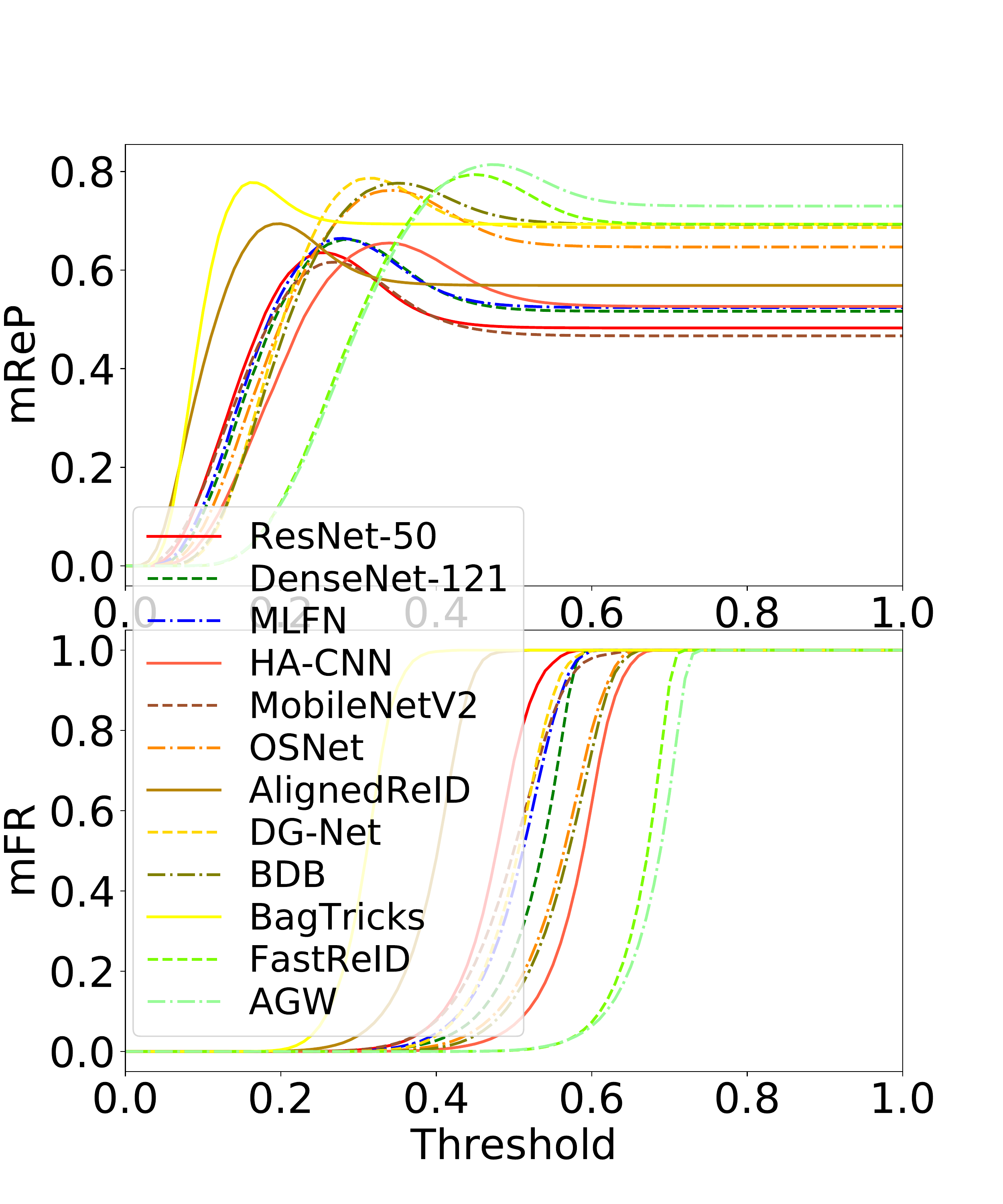} &
		\includegraphics[width=.249\textwidth]{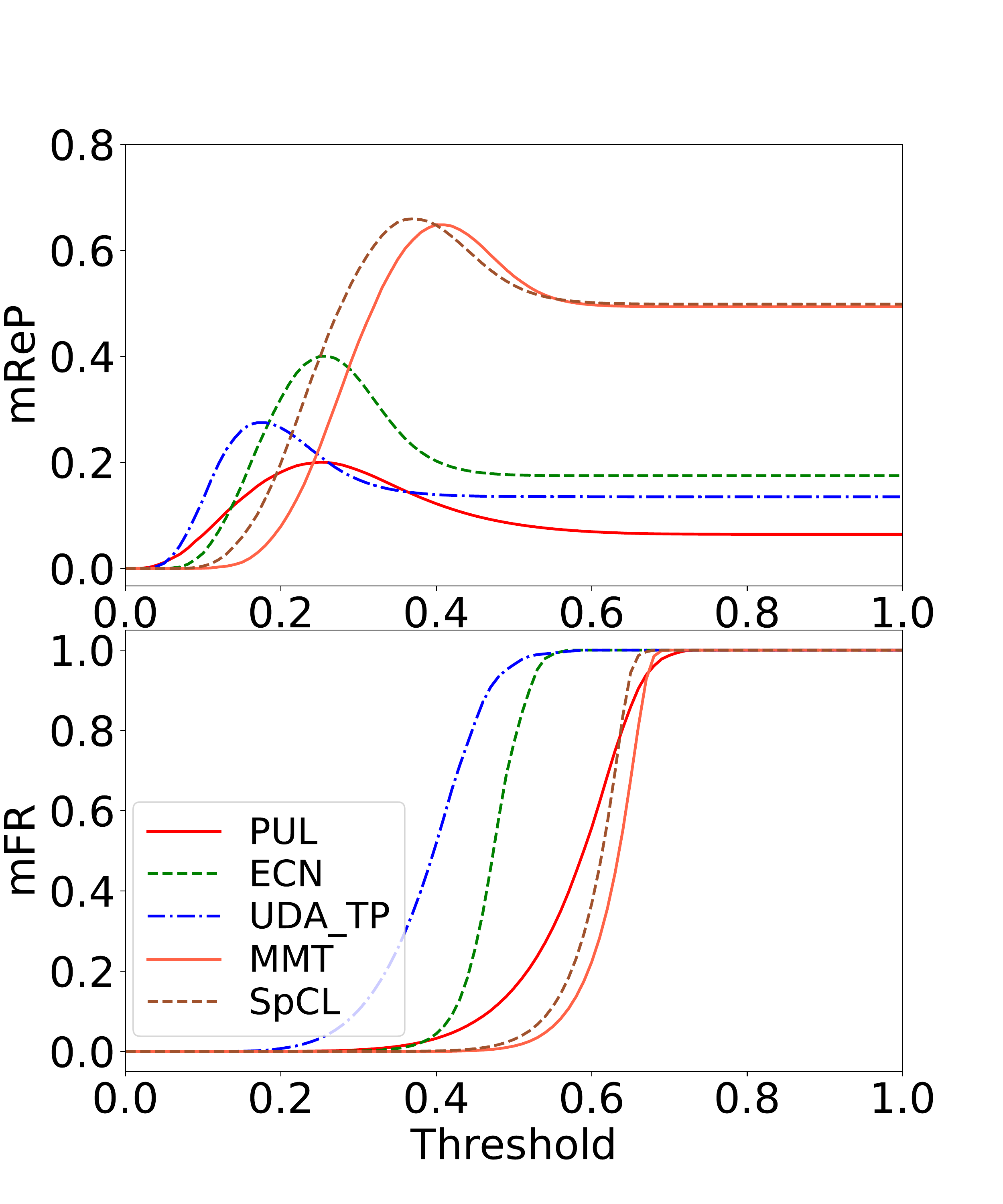} &
        \includegraphics[width=.249\textwidth]{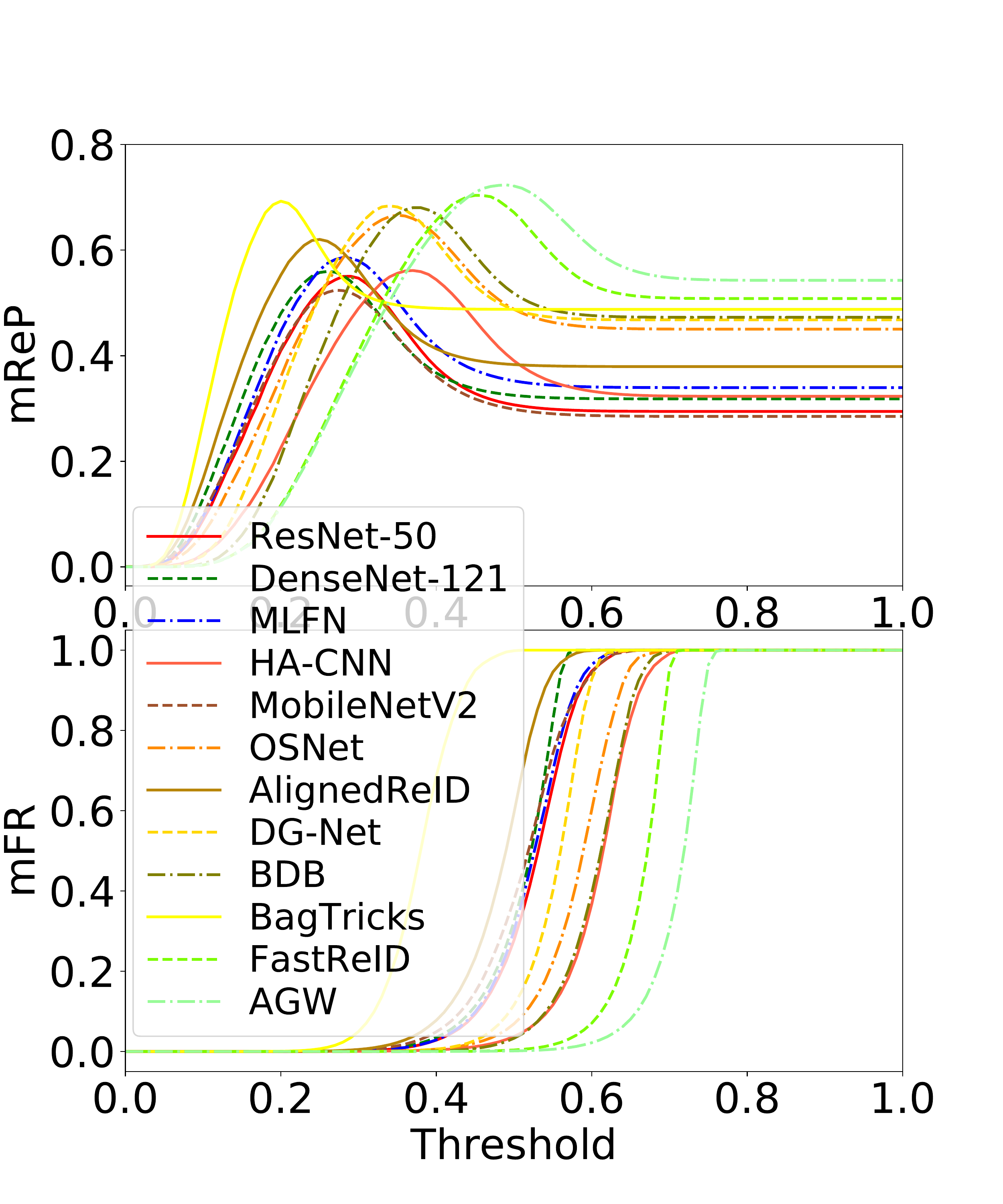} & 
		\includegraphics[width=.249\textwidth]{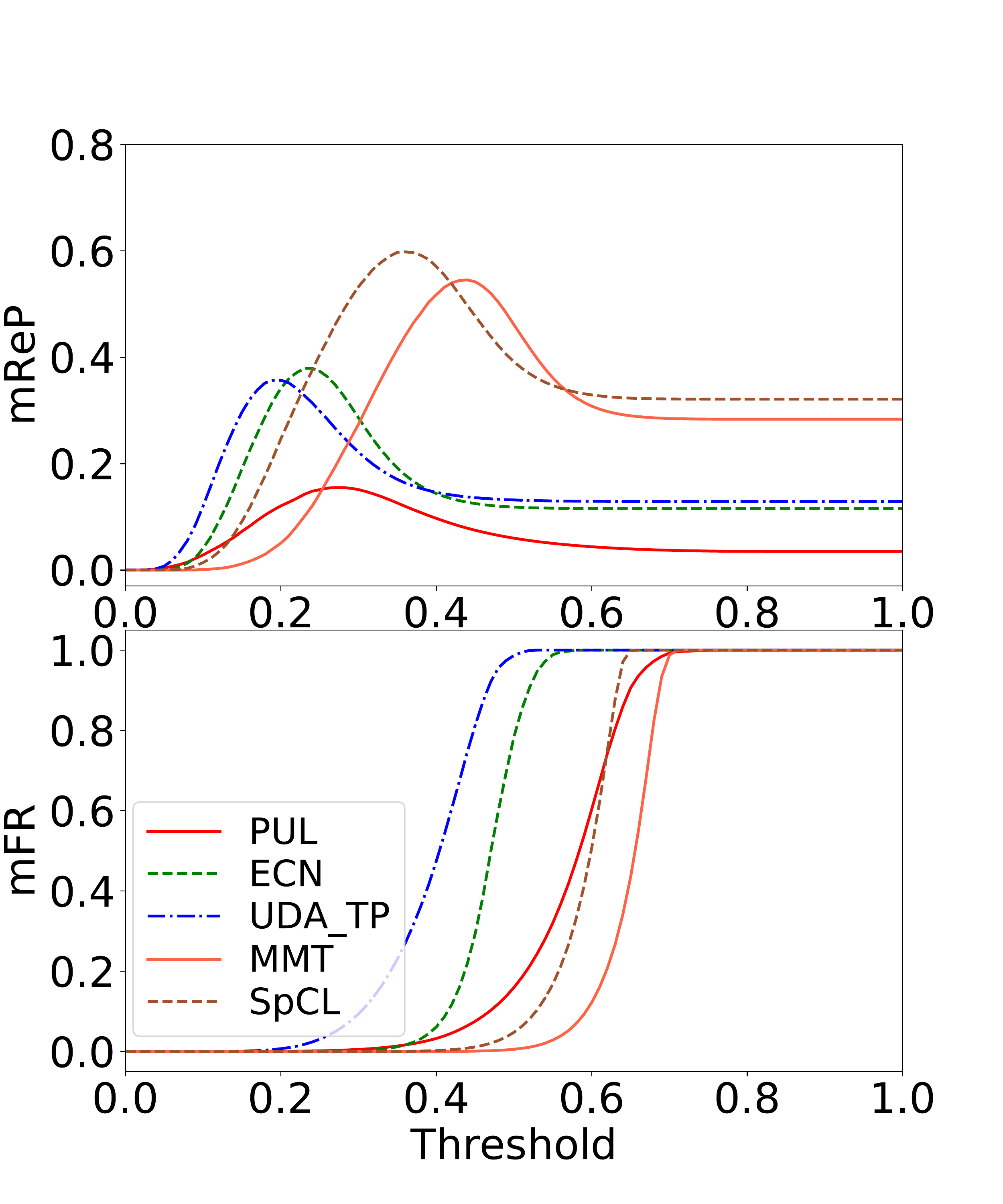} \\
		(a)  & (b) & (c)  & (d) \\
	\end{tabular}
\caption{The $GOM$ results, \textit{i.e.}, $mReP$ and $mFR$ curves (\%) in terms of threshold on the person re-ID datasets. (a) and (c) are the results of supervised methods on the Market-1501 and DukeMTMC-reID datasets, respectively; (b) and (d) are the results of unsupervised methods on the Market-1501 and DukeMTMC-reID datasets, respectively.}
\label{fig:person-sota}
\end{figure*}

\tabcolsep=4pt
\begin{table*}[t]
\centering
\caption{The comparison results on the person re-ID datasets. The results by existing metrics CMC@1 (\%) and mAP (\%), as well as our metrics $mVP_{max}$ (\%), $mReP_{max}$ (\%), $MREP$ (\%), and $MFR$ (\%) are reported.}
\resizebox{\linewidth}{!}{
\begin{tabular}{l l| cc|cccc | cc|cccc}
\toprule
& \multirow{3}{*}{Methods} & \multicolumn{6}{c|}{Market-1501} & \multicolumn{6}{c}{DukeMTMC-reID}            \\
\cline{3-14} 
& & \multicolumn{2}{c|}{Existing} & \multicolumn{4}{c|}{Proposed}& \multicolumn{2}{c|}{Existing}  &      \multicolumn{4}{c}{Proposed}     \\ 
&        &CMC@1 &mAP  &$mVP_{max}$ &$mReP_{max}$ &$MREP$ &$MFR$$\downarrow$ 
         &CMC@1 &mAP  &$mVP_{max}$ &$mReP_{max}$ &$MREP$ &$MFR$$\downarrow$ \\ 
\midrule
\midrule
\multirow{12}{*}{Supervised} & MobileNetV2~\cite{sandler2018mobilenetv2} 
                                & 86.8 & 69.9 & 48.9 & 61.6 \yc{$_{0.27}$} & 43.7 & 50.7
                                & 76.6 & 57.6 & 39.9 & 52.4 \yc{$_{0.27}$} & 29.7 & 48.9 \\ 
& ResNet-50~\cite{he2016deep}   & 88.2 & 71.8 & 51.2 & 63.5 \yc{$_{0.26}$} & 44.9 & 52.8 
                                & 78.0 & 60.2 & 42.5 & 55.0 \yc{$_{0.29}$} & 30.5 & 47.6 \\
& MLFN~\cite{chang2018multi}    & 89.7 & 74.7 & 54.6 & 66.4 \yc{$_{0.28}$} & 47.7 & 49.7 
                                & 80.9 & 64.2 & 46.3 & 58.6 \yc{$_{0.28}$} & 34.2 & 48.0 \\ 
& DenseNet-121~\cite{huang2017densely}  
                                & 90.2 & 74.8 & 54.1 & 66.2 \yc{$_{0.28}$} & 47.0 & 47.7 
                                & 78.5 & 61.6 & 43.6 & 56.0 \yc{$_{0.26}$} & 32.5 & 49.1 \\ 
& HA-CNN~\cite{li2018harmonious}& 90.6 & 75.3 & 53.5 & 65.5 \yc{$_{0.34}$} & 46.3 & 42.0 
                                & 79.3 & 62.9 & 43.6 & 56.1 \yc{$_{0.37}$} & 31.4 & 39.3 \\ 

& AlignedReID~\cite{zhang2017alignedreid}  
                               & 90.6  & 77.7 & 58.4 & 69.4 \yc{$_{0.20}$} & 53.7 & 60.7 
                               & 81.2  & 67.4 & 50.4 & 62.0 \yc{$_{0.25}$} & 37.8 & 51.9 \\  
& OSNet~\cite{zhou2019omni}    & 94.2  & 83.6 & 66.6 & 76.2 \yc{$_{0.35}$} & 56.2 & 44.6 
                               & 87.0  & 72.7 & 55.4 & 66.6 \yc{$_{0.35}$} & 41.9 & 41.9 \\ 
& BDB~\cite{dai2019batch}      & 94.2  & 85.6 & 68.9 & 77.6 \yc{$_{0.35}$} & 58.2 & 43.9 
                               & 88.0  & 75.5 & 56.7 & 68.0 \yc{$_{0.37}$} & 41.0 & 39.6 \\ 
                               
& BagTricks~\cite{luo2019bag}  & 93.7  & 85.8 & 68.9 & 77.8 \yc{$_{0.16}$} & 64.2 & 69.3 
                               & 86.8  & 75.3 & 58.8 & 69.3 \yc{$_{0.20}$} & 46.4 & 62.2 \\ 
& DG-Net~\cite{zheng2019joint} & 94.8  & 86.1 & 69.6 & 78.6 \yc{$_{0.32}$} & 58.1 & 50.2 
                               & 86.0  & 74.9 & 57.5 & 68.4 \yc{$_{0.34}$} & 42.0 & 44.9 \\ 
& FastReID~\cite{hefastreid}   &{94.3} &{86.5} & 70.8 & 79.4 \yc{$_{0.45}$} & 52.9 & 33.8 
                               &{86.8} &{77.0} & 60.2 & 70.4 \yc{$_{0.45}$} & 41.7 & 33.9  \\ 
& AGW~\cite{ye2020deep}        &{95.5} &{88.3} &{73.4} &{81.4} \yc{$_{0.47}$} & 54.8 & {32.3} 
                               &{89.2} &{79.6} &{62.4} &{72.3} \yc{$_{0.49}$}  & 43.8 & {29.2} \\ 
\midrule
\multirow{5}{*}{Unsupervised} & PUL~\cite{fan18unsupervisedreid}  
                               & 41.9 & 19.2 & 11.3 & 20.0 \yc{$_{0.25}$} & 9.3  & 42.5 
                               & 27.3   & 15.2   & 9.6 & 15.5 \yc{$_{0.27}$}  & 6.2  & 43.0  \\ 
& UDA\_TP~\cite{song2020unsupervised}  & 58.6   & 34.7   & 16.8 & 27.5 \yc{$_{0.18}$}  & 14.5  & 61.0 
                                       & 64.4   & 42.3   & 24.1 & 35.7 \yc{$_{0.20}$} & 15.6  & 60.7 \\ 
& ECN~\cite{zhong2019invariance} & 76.6   & 44.5   & 25.4 & 40.1\yc{$_{0.26}$}  & 18.8  & 52.9  
                               & 64.0   & 41.2   & 25.6 & 37.9 \yc{$_{0.24}$}  & 14.6  & 53.2 \\ 
& MMT~\cite{ge2020mutual}      &88.1 &74.3 &52.8 &64.9 \yc{$_{0.40}$} &39.3  &{37.5} 
                               &75.6 &60.3 &42.6 &54.5 \yc{$_{0.44}$} &26.1  &{35.4} \\ 
& SpCL~\cite{ge2020selfpaced}  &{89.5} &{76.0} &{54.1} &{66.0} \yc{$_{0.37}$} &{42.0}  &39.7 
                               &{82.4} &{67.1} &{47.7} &{59.8} \yc{$_{0.36}$} &{32.1}  &41.2 \\ 
\bottomrule
\end{tabular}}
\label{tb:person-sota}
\end{table*}

\tabcolsep=4pt
\begin{table}[t]
\centering
\caption{Results (\%) on the vehicle re-ID dataset.}
\resizebox{\linewidth}{!}{
\begin{tabular}{l| cc|cccc}
\toprule
\multirow{3}{*}{Methods} & \multicolumn{6}{c}{VeRi776}  \\ 
\cline{2-7} 
& \multicolumn{2}{c|}{Existing}  &  \multicolumn{4}{c}{Proposed}  \\ 
& CMC@1 &mAP  & $mVP_{max}$ &$mReP_{max}$ & $MREP$ & $MFR$ $\downarrow$ \\ 
\midrule
\midrule
ResNet-50~\cite{he2016deep}   & 92.3 & 65.0 & 42.7 & 58.1 \yc{$_{0.32}$} & 34.9 & 46.6 \\ 
VOC-ReID~\cite{Zhu_2020_CVPR_Workshops}    & 95.2 & 77.0 & 55.2 & 68.8 \yc{$_{0.43}$} & 41.6 & 39.2  \\
BagTricks~\cite{luo2019bag} & 95.0 & 79.8 & 55.9 & 68.9 \yc{$_{0.35}$} & 48.4 & 41.0\\ 
\bottomrule
\end{tabular}}
\label{tb:v-sota}
\end{table}

\section{Experimental Validation}
Sec.~\ref{Sec: analysis of GOM} provides a theoretical analysis of the re-ID errors, and compares $GOM$ with existing metrics via toy samples. In this section, we verify the effectiveness of $GOM$ in the actual application task. First, we conduct qualitative and quantitative experimental evaluations on person and vehicle re-ID tasks. Second, we verify the effectiveness and evaluation quality from the level of human visual assessment study. 

\subsection{Evaluating re-ID benchmarks with GOM}
\label{Sec:evaluating GOM}

\heading{Experimental settings.}
We assess $GOM$ on two person re-ID datasets (\textbf{Market-1501}~\cite{zheng2015scalable} and \textbf{DukeMTMC-reID}~\cite{Zheng_2017_ICCV}) and one vehicle re-ID dataset (\textbf{VeRi776}~\cite{Liu_2016_ECCV}). To simulate the open-set re-ID setting, we randomly select 100 images from other datasets as new queries. In person re-ID evaluations, we compare a batch of state-of-the-art methods, both supervised and unsupervised. Especially,  we discuss the performance with $GOM$ under different thresholds and give some explorative solutions. We restrict our evaluation to only those methods that are published in peer-reviewed journals and conferences. We evaluate 12 supervised methods~\cite{he2016deep,chang2018multi,huang2017densely,li2018harmonious,zhang2017alignedreid,zhou2019omni,dai2019batch,sandler2018mobilenetv2,luo2019bag,zheng2019joint,ye2020deep,hefastreid} and 5 unsupervised methods~\cite{fan18unsupervisedreid,zhong2019invariance,song2020unsupervised,ge2020mutual,ge2020selfpaced} on Market-1501 and DukeMTMC-reID, and 3 vehicle re-ID methods~\cite{he2016deep,Zhu_2020_CVPR_Workshops,luo2019bag} on VeRi776.

\heading{Datasets.} Market-1501~\cite{zheng2015scalable} is a large scale person re-ID dataset, containing 32,688 labeled images, 12,936 images of 751 identities for training, and 19,732 images of 750 identities for testing. DukeMTMC-reID~\cite{Zheng_2017_ICCV} is also a popular evaluated person re-ID dataset, which has 16,522 training images, 2,228 query images, and 17,661 gallery images. VeRi776~\cite{Liu_2016_ECCV} is a typical vehicle re-ID dataset, which consists of over 50,000 images of 776 vehicles.

\heading{Re-ID results by GOM.} Figure~\ref{fig:person-sota} shows the $mReP$ and $mFR$ curves of all evaluated methods on Market-1501 and DukeMTMC-reID datasets. In addition to the results by existing metrics CMC@1 and mAP, Table~\ref{tb:person-sota} shows the results of $mVP_{max}$, $mReP_{max}$, $MREP$, and $MFR$ on two person re-ID datasets.  Table~\ref{tb:v-sota} shows the results on the vehicle re-ID dataset. The detailed analysis is as follows:

1) Figure~\ref{fig:person-sota} shows that the curves of all methods have highly consistent trends and shapes on two different datasets. The pros and cons of different methods on two datasets are also extremely similar. These demonstrate the good universality of our metric.

2) Table~\ref{tb:person-sota} shows that the method has excellent performance evaluated by retrieval based metrics (mAP), whereas it may not achieve good performance evaluated by verification based metrics ($mVP_{max}$). For example, evaluated on the Market-1501 dataset, DenseNet-121 has a better mAP score than MLFN does, but it does not have a better $mVP_{max}$ score. It shows that our metric introduce more focus on the different aspects of evaluated methods.

3) Figure~\ref{fig:person-sota} shows that different methods touch the top $mReP_{max}$ scores at different thresholds. It means that when we want to adopt a method to some applications, we should set a proper threshold to obtain its best performances. Here, we list the threshold $\tau_{max}$ along with the $mReP_{max}$ in Table~\ref{tb:person-sota}.

4) Table~\ref{tb:person-sota} shows that the supervised method AGW and unsupervised method SpCL obtain the best $mReP_{max}$ scores on both person re-ID datasets, respectively. A high $mReP_{max}$ score indicates that within the returned samples, high ranked samples are very likely to be the targets. $mReP_{max}$ demonstrates that it has powerful competitiveness in the verification task and ensures the reliability of the retrieval.

5) Both Table~\ref{tb:person-sota} and Table~\ref{tb:v-sota} show that the pros and cons of results among different methods are similar by existing metrics and our metrics. Since the retrieval has a very strong relationship with verification, and also contributes a lot to the GOM metric. Thus, it is reasonable that existing methods and proposed GOM have high global similarity, and some local variations.

6) The BagTricks method gets the highest $MREP$ score. As discussed above, to get a high $mReP_{max}$ score, we need to know the best threshold of the selected method. If we do not know this threshold, BagTricks is the best choice.

7) The BagTricks method also gets the highest $MFR$ score. Thus, although it performs very well in the closed-world re-ID, it does not fit the open-set re-ID well. Note that existing metrics cannot provide this suggestion.

\heading{Summary.} 1) Most of the current methods have little overall difference. If an analysis is made from the perspectives of re-ID, the performance evaluation cannot be made from the value of the mAP alone. Our GOM can provide comprehensive performance evaluation from an intuitive perspective ($mReP$ and $mFR$ curves) and objective perspective ($mVP_{max}$, $mReP_{max}$, $MREP$, and $MFR$). 2) Different sub-metrics show the importance of different aspects, for instance, if the verification is the most important for an application, a method with the best $mVP_{max}$ should be selected; if verification and retrieval are both considered and balanced, a method with the best $mReP_{max}$ should be selected (its threshold should be recorded); if the threshold cannot be set fixed, a method with the best $MREP$ should be selected; if we need to consider the open-set problem, and do not like so many noise influence the application, a method with the best $MFR$ should be selected.

\subsection{Human visual assessment study}
\label{Sec: Human Visual Assessment Study}
In this subsection, we perform a user study in order to determine how designed metrics align with the human judgment of re-ID quality. We invite 13 participants to conduct a user study. Each participant is asked to review the results of 21 hard queries generated by five typical re-ID methods on the Market-1501 dataset. The results have two forms, one is a ranking list for each query, the other is a batch of returned samples by a threshold of $mReP_{max}$, which are all considered as the targets. During browsing the ranking list, the participant search out the targets in each ranking list, and the cost time is recorded. After reviewing each batch of returned samples, the participant gives a rating to five evaluated methods (1-5, 5 is the highest score to indicate the best method). Finally, we have the average ratings for different methods and the average time cost of searching out targets in each ranking list. The results are recorded in Table~\ref{tb:user}. Since we consider that re-ID quality consists of quality of verification and quality of retrieval. Our $mReP_{max}$ and $MREP$ sub-metrics both consider quality verification and retrieval. For the user part, our average ratings and average search time reflect the quality verification and retrieval, respectively. Thus, Table~\ref{tb:user} shows that the designed metrics well align with the human judgment of re-ID quality.

\tabcolsep=4pt
\begin{table}[t]
\centering
\caption{The results of user study, we report the results of mAP, $mReP_{max}$, $MREP$, average rating (1-5), and average search time (mins) on five different methods. }
\resizebox{\linewidth}{!}{
\begin{tabular}{l| ccccc}
\toprule
Methods  & mAP  &$mReP_{max}$ & $MREP$ & avg Rating & avg Time $\downarrow$\\ 
\midrule
\midrule
MobileNetV2~\cite{sandler2018mobilenetv2}    & 69.9 & 61.6 & 44.7 & 2.26 & 5.55 \\ 
HA-CNN~\cite{li2018harmonious}               & 75.3 & 65.5 & 47.4 & 2.37 & 4.34\\ 
AlignedReID~\cite{zhang2017alignedreid}      & 77.7 & 69.4 & 55.1 & 3.53 & 3.47\\ 
FastReID~\cite{hefastreid}                   & 86.5 & 79.4 & 53.9 & 4.12 & 3.13\\ 
AGW~\cite{ye2020deep}                        & 88.3 & 81.4 & 55.7 & 4.32 & 1.95\\ 
\bottomrule
\end{tabular}}
\label{tb:user}
\end{table}

\section{Conclusion}
In this paper, we analyzed the current re-ID metrics based on three types of errors and showed their limitations. Then a new metric, Genuine Open-set re-ID Metric (GOM), has been proposed for the re-ID task. GOM shows the comprehensive performance of re-ID methods from the respective verification task and retrieval task. We consider that existing metrics are designed for retrieval task, essentially, not suitable for re-ID. With GOM, researchers can evaluate re-ID methods in a thorough way, and developers are able to select suitable algorithms to build their application. We encourage the re-ID community to consider this metric in future algorithms evaluations and comparisons.

\heading{Prospects.}
Note that even our proposed GOM metric is comprehensive, there are still some directions worth noting. 1) All existing methods cannot evaluate a dynamic gallery system or an online application. It is because the gallery size and the outputs always change. Thus, designing a metric for the dynamic gallery system is a prospective direction. 2) We still didn't consider the situations that the results are separated, fragmented, or even protected due to the privacy concern. How to design a metric to address these issues is also a direction that should be discussed. In the future, we will continuously investigate how to make a more comprehensive evaluation for re-ID.

{\small
\bibliographystyle{ieee_fullname}
\bibliography{egbib}
}
\end{document}

%% file: setup/macro.tex
\newcommand{\yc}[1]{\textcolor{xgray}{#1}}

\renewcommand\fbox{\fcolorbox{blue!20}{white}}

\def\ie{\textit{i.e.,}~}

\def\etal{\textit{et~al.}~}

\newcommand{\heading}[1]{\noindent\textbf{#1}}

\newcolumntype{x}[1]{>{\centering\arraybackslash}p{#1pt}}
\newcolumntype{y}[1]{>{\raggedright\arraybackslash}p{#1pt}}
\newcolumntype{z}[1]{>{\raggedleft\arraybackslash}p{#1pt}}

\newlength\savewidth

\newcommand{\ignore}[1]{}   

%% file: setup/color.tex
\usepackage{xcolor}
\colorlet{dark-blue}{blue!50!black}
\colorlet{dark-cyan}{cyan!75!black}
\colorlet{dark-purple}{purple!50!black}
\colorlet{dark-red}{red!75!black}
\colorlet{dark-green}{green!75!black}
\colorlet{dark-orange}{orange!50!black}
\colorlet{dark-gray}{black!75}
\colorlet{light-gray}{black!30}
\definecolor{nice-red}{HTML}{E41A1C}
\definecolor{nice-orange}{HTML}{FF7F00}
\definecolor{nice-yellow}{HTML}{FFC020}
\definecolor{nice-green}{HTML}{39b54a}
\definecolor{nice-blue}{HTML}{0071bc}
\definecolor{nice-purple}{HTML}{984EA3}

\definecolor{darkGreen}{rgb}{0, 0.6, 0}
\definecolor{darkRed}{rgb}{0.9, 0, 0} 
\definecolor{cyan}{rgb}{0, 0.5, 0.6} 
\definecolor{darkViolet}{rgb}{0.58, 0, 0.83}

\definecolor{lightgreen}{rgb}{0.4, .9, 0.4}
\definecolor{lightred}{rgb}{1, 0.5, 0.51}
\definecolor{xgray}{rgb}{0.6, 0.6, 0.6}
\definecolor{Highlight}{HTML}{39b54a}
\definecolor{citecolor}{HTML}{0071bc}